# Graph neural network-based surrogate modelling for real-time hydraulic prediction of urban drainage networks

Zhiyu Zhang[1,2], Chenkaixiang Lu[3], Wenchong Tian[1,4], Zhenliang Liao[1,2,3*], Zhiguo Yuan[5]

1 College of Environmental Science and Engineering, Tongji University, 200092, Shanghai, China.

2 Key Laboratory of Yangtze River Water Environment, Ministry of Education, Tongji University, 200092, Shanghai, China.

3 College of Civil Engineering and Architecture, Xinjiang University, 830046, Urumqi, China.

4 Key Laboratory of Urban Water Supply, Water Saving and Water Environment Governance in the Yangtze River Delta of Ministry of Water Resources, 200092, Shanghai, China.

5 School of Energy and Environment, City University of Hong Kong, Hong Kong SAR, China

* Corresponding author: Zhenliang Liao (zl_liao@tongji.edu.cn)

**Abstract**

Physics-based models are computationally time-consuming and infeasible for real-time scenarios of urban drainage networks, and a surrogate model is needed to accelerate the online predictive modelling. Fully-connected neural networks (NNs) are potential surrogate models, but may suffer from low interpretability and efficiency in fitting complex targets. Owing to the state-of-the-art modelling power of graph neural networks (GNNs) and their match with urban drainage networks in the graph structure, this work proposes a GNN-based surrogate of the flow routing model for the hydraulic prediction problem of drainage networks, which regards recent hydraulic states as initial conditions, and future runoff and control policy as boundary conditions. To incorporate hydraulic constraints and physical relationships into drainage modelling, physics-guided mechanisms are designed on top of the surrogate model to restrict the prediction variables with flow balance and flooding occurrence constraints. According to case results in a stormwater network, the GNN-based model is more cost-effective with better hydraulic prediction accuracy than the NN-based model after equal training epochs, and the designed mechanisms further limit prediction errors with interpretable domain knowledge. As the model structure adheres to the flow routing mechanisms and hydraulic constraints in urban drainage networks, it provides an interpretable and effective solution for data-driven surrogate modelling. Simultaneously, the surrogate model accelerates the predictive modelling of urban drainage networks for real-time use compared with the physics-based model.

**Graphical Abstract**

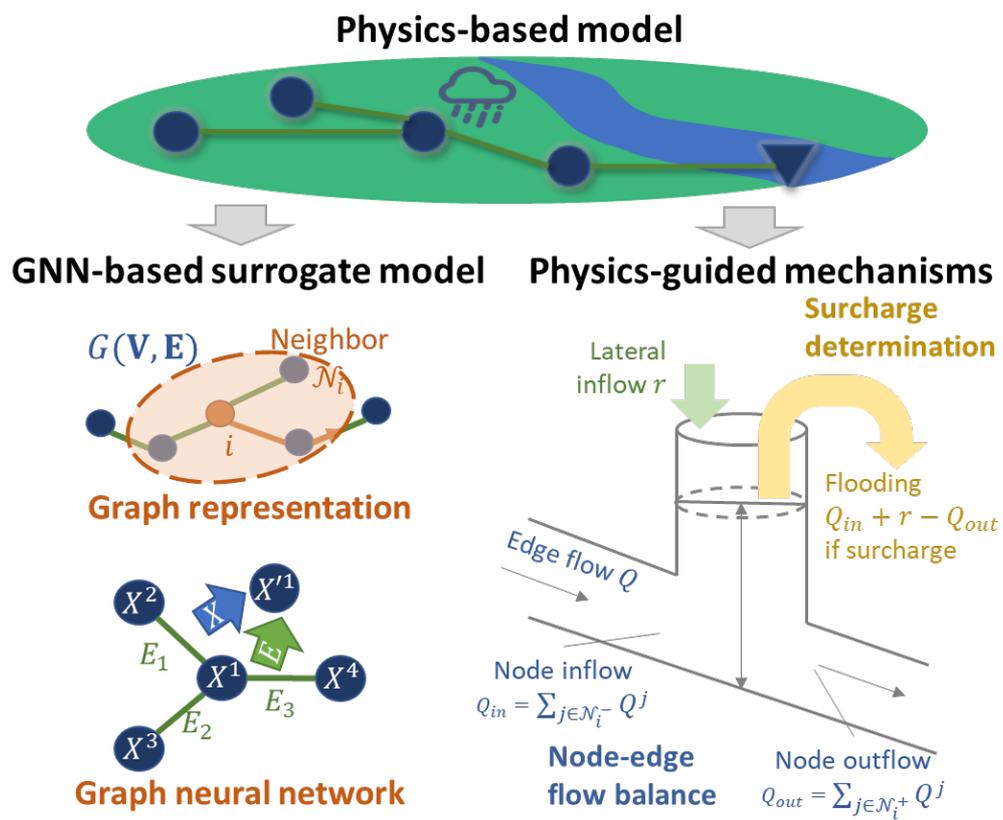

## 1. Introduction

Urban drainage modelling plays a key role in the management of drainage systems, and enables detailed simulation and analysis of infrastructures in sewage collection and storm drainage scenarios from design to monitoring and operational stages (Fletcher et al., 2013; Garcia et al., 2015; Garzón et al., 2022). Physics-based modelling approaches follow hydraulic disciplines to solve Saint-Venant equations by numerical methods, which is time-consuming and computationally expensive for application scenarios in need of repetitive simulation-optimizations or an instant response such as real-time control (van Daal-Rombouts et al., 2016). Owing to powerful nonlinear fitting abilities, machine learning surrogate models (MLSMs) accelerate simulation for practical use such as flooding (Li et al., 2023) and flow (Sufi Karimi et al., 2019) prediction, real-time control (Luo et al., 2023; Tian et al., 2022), hydraulic simulation (Palmitessa et al., 2022), and optimization of green infrastructures (Seyedashraf et al., 2021). Compared with physics-based models, MLSMs split induction and inference into training and application stages, in which a trained model only conducts a fast feedforward calculation for emulation, instead of solving complex partial differential equations. In addition, deep learning models make use of graph processing units (GPU) for data parallelization, which further accelerates emulations of multiple scenarios in a batch (Fu et al., 2022).

A flow routing model consists of state variables such as water depth and flow rate of each manhole and pipe, governed by complex differential equations. However, most MLSMs in existing studies are designed to fit a specific variable (e.g., the objective function of optimization) (Li et al., 2022b; Luo et al., 2023; Tian et al., 2022), which results in a "black-box" model that does not demonstrate the mechanisms. A residual neural network recursively simulates full hydrodynamics of flow and depth variables in urban drainage networks (Palmitessa et al., 2022). However, the fully-connected model structure is inefficient, with redundant neural connections of variables for all the manholes and pipes. This is especially the case in large drainage systems, as there is little correlation between hydraulic variables at distant catchments in a short horizon. There is a research gap on how to design an interpretable and efficient MLSM to emulate complex and nonlinear hydraulic dynamics and predict future state transitions for real-time use.

As an inductive bias of deep learning, graph neural networks (GNNs) have shown expressive power in processing non-Euclidean data structures for node and link prediction tasks (Garzón et al., 2022; Zhou et al., 2020). GNN appears to be a suitable modelling framework for urban water networks due to their inherent graph structure (Fu et al., 2022), and existing literature has contributed to nodal pressure reconstruction (Hajgató et al., 2021), state estimation (Xing and Sela, 2022), and water quality prediction (Li et al., 2024) in drinking water distribution networks.

Inspired by the graph structure of urban drainage networks, we hypothesize GNN can be used in the surrogate model of flow routing process along the network for hydraulic predictions. Hydraulic variables such as water level and flow rate are attached to manholes and pipes in an urban drainage network, which can be represented as node and edge features in a graph. Drainage flow routing follows the topological structure from upstream to downstream, and hydraulic state transitions are highly related to neighbors in the drainage network. Therefore, GNN may provide an effective solution considering topological information to emulate the flow routing along drainage networks, which leverages the existing relationship as an inductive bias to reduce the difficulty of training the surrogate model (Garzón et al., 2022).

When full hydraulic states of manholes and pipes are simulated with a physics-based model, the flow balance is considered at each element to ensure numerical convergence. Such an approach can also be adopted in surrogate modelling to limit the water balance errors of flow predictions using physics-guided mechanisms, especially for flooding estimation which plays a key role in evaluating drainage conditions. Therefore, the output flow variables of the element-level surrogate model can be regulated with hydraulic constraints, which will likely enhance the modelling performance with given physical laws as "shortcuts", and also preserve the interpretability of data-driven modelling in urban drainage networks (Palmitessa et al., 2022).

The main aim of this work is to provide a surrogate model for element-level hydraulic prediction of urban drainage networks to accelerate the predictive modelling process for real-time use, and to incorporate physical constraints to enhance prediction accuracy with domain knowledge. The hydraulic prediction problem requires a model to simulate flow routing state variations of both node (e.g., manholes and tanks) and edge (e.g., conduits and orifices) elements, with updated hydraulic states as initial conditions, and future runoff inflow and control settings as boundary conditions. This work develops a spatio-temporal surrogate model based on GNN (Farahmand et al., 2023; Jin et al., 2023a; Yuan et al., 2022), with a fully-connected neural network (NN) as a reference. Considering flow balance in urban drainage hydraulics, we designed two physics-guided mechanisms to limit prediction errors. These included a node-edge fusion modelling approach that incorporates the flow relationship between manholes and their connected pipes, and a flooding determination technique that calculates flooding only when the surcharge is determined. To the knowledge of the authors, this is the first attempt to use GNN in surrogate modelling of urban drainage networks.

## 2. Methodology

A surrogate model is proposed and trained to emulate an existing physics-based flow routing model (by fitting the data generated by the physics-based model) for real-time prediction of hydraulic routing variables. Section 2.1 introduces the hydraulic prediction problem. Section 2.2 provides the general structure of the surrogate model. Section 2.3 elaborates on different spatial convolution blocks based on the fully-connected neural network (NN) and the GNN. Section 2.4 proposes two physics-guided mechanisms as hydraulic constraints in the surrogate model. Section 2.5, 2.6, and 2.7 provide training, evaluation metrics and technical implementation details.

### 2.1 Hydraulic prediction problem

A hydraulic prediction problem of urban drainage networks is defined as follows. As Figure 1(a), the hydraulic states of a urban drainage network consist of variables $X_t, E_t$ for all the elements (nodes and edges respectively) at time step $t$, including water depth $h_{v,t}^i$, inflow from upstream pipes $Q_{in,t}^i$ and outflow to downstream pipes $Q_{out,t}^i$ for each node $i$, and water depth $h_{e,t}^j$, and flow $Q_t^j$ for each link $j$. The lateral inflow $r_t^i$ (including runoff and dry weather inflow) is regarded as a given constant for each node $i$ as the rainfall-runoff is considered as the boundary. The control setting $a_t^j$ of edge $j$ equals the opening percentage for a controllable asset and equals 1 for a conduit. $Q_{w,t}^i$ is the flooding overflow of node $i$. The hydraulic prediction problem is designed to predict future hydraulic states (with potential flooding) $\{X_{t+u}, E_{t+u}, Q_{w,t+u} | u = 1, \dots, n\}$, given previous states $\{X_{t-u}, E_{t-u}, R_{t-u}, A_{t-u} | u = 0, \dots, m-1\}$ and future boundary conditions $\{R_{t+u}, A_{t+u} | u = 1, \dots, n\}$, as depicted in Figure 1(b). A one-minute time step is used for each variable, and all the flow variables refer to the volume flow in each time step $(m^3/min)$.

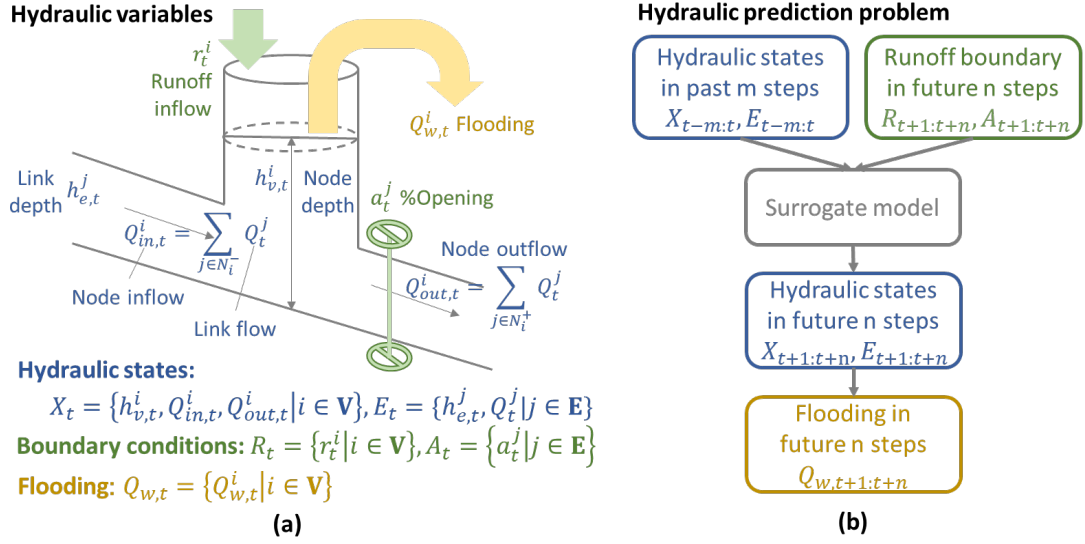

Figure 1. Hydraulic variable definitions and representations of nodes and edges. $X_t, E_t$ are the state variables of nodes and edges respectively. $R_t, A_t$ represent the runoff inflow of nodes and control settings of edges. $\mathbf{V}, \mathbf{E}$ are sets of nodes and edges in the drainage network. $\mathcal{N}_i^-$ and $\mathcal{N}_i^+$ are directly connected inflow and outflow edge sets of node $i$.

## 2.2 Surrogate modelling structure

A surrogate model is proposed for the spatio-temporal representations of the flow routing hydrodynamics in drainage networks. The model mainly consists of 2 spatial convolution blocks, 2 temporal blocks, and a constraint block as shown in Figure 2. The model has a $m$-step state input, a $n$-step boundary input and a $n$-step state output, and temporal convolutional layers are adopted in 2 temporal blocks to extract the asynchronous hydrodynamics in both past and future domain (van den Oord et al., 2016; Yu and Koltun, 2015). A skipped connection is used for initializing hydraulic states and using neural networks to learn the residual similar to Palmitessa et al. (2022).

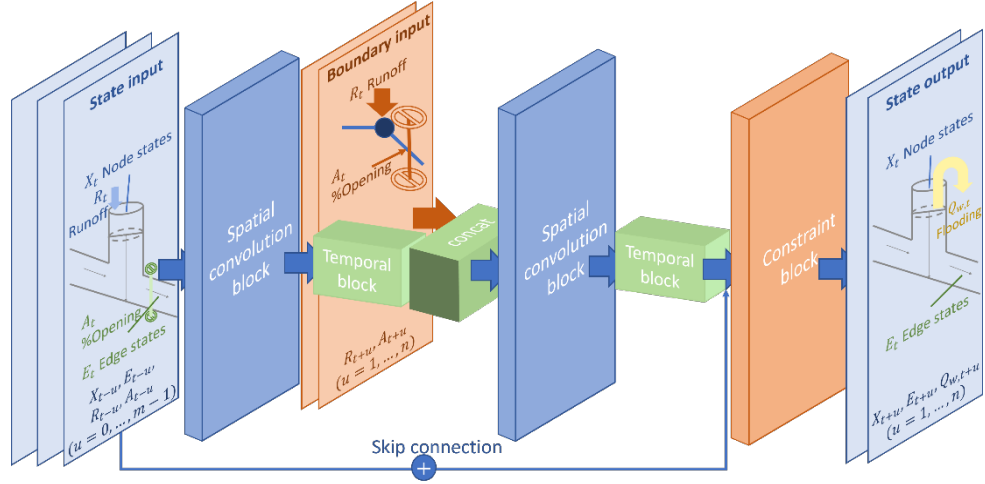

Figure 2. The spatio-temporal structure of the surrogate model for hydraulic state prediction. Note that $m, n$ are the number of past and future time steps.

## 2.3 Spatial convolution block

The spatial convolution block emulates flow routing along the drainage network for spatial correlations of variables with a fully-connected NN or a GNN, which differ in hydraulic variable representations (Figure 3(a)(c)) and convolution layers (Figure 3(b)(d)) and are mainly compared and investigated in this work.

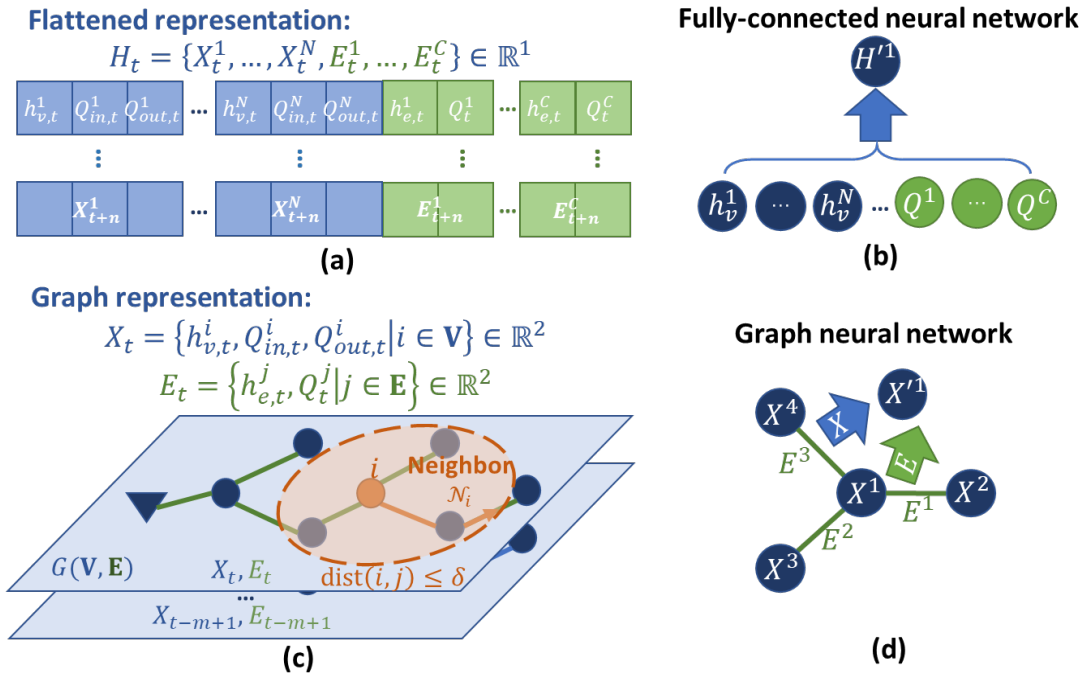

Figure 3. Hydraulic variable representations in the fully-connected neural network (a) and the graph neural network (c), and spatial convolution layers in the fully-connected neural network (b) and the graph neural network (d). $H_t$ is the flattened representation

of both node and edge variables. $\mathbf{V}, \mathbf{E}$ and $N, C$ are the sets and numbers of nodes and edges respectively.

### 2.3.1 Fully-connected neural network

As illustrated in Figure 3(a), the hydraulic variables of all the elements in the drainage network can be flattened into a total number of $3N + 2C$ variables in each time step and represented as $H_t = \{h_{v,t}^1, Q_{in,t}^1, Q_{out,t}^1, \dots, h_{v,t}^N, Q_{in,t}^N, Q_{out,t}^N, h_{e,t}^1, Q_t^1, \dots, h_{e,t}^C, Q_t^C\}$ (including all the hydraulic variables in Figure 1(a)), which is an one-dimensional vector in time step $t$ and enables a fully-connected NN to establish their global relationship as Eq. 1 and Figure 3(b) (Palmitessa et al., 2022),

$$H' = \sigma(\boldsymbol{W}H + \boldsymbol{b}) \tag{1}$$

where $H, H'$ are the input and output hidden features (for all the time step) of a NN layer. $\boldsymbol{W}$ is the weight matrix, $\sigma$ is the activation function, and $\boldsymbol{b}$ is the bias.

### 2.3.2 Graph neural network

The urban drainage network can also be regarded as a graph $G(\mathbf{V}, \mathbf{E})$ following the drainage topology connections to represent all the element-level hydraulic variables into individual nodes $\mathbf{V}$ and edges $\mathbf{E}$ of the graph structure as $X_t = \{h_{v,t}^i, Q_{in,t}^i, Q_{out,t}^i | i \in \mathbf{V}\}, E_t = \{h_{e,t}^j, Q_t^j | j \in \mathbf{E}\}$, which have both spatial and variable dimensions and are suitable as the input of a graph neural network (GNN). The undirected graph type is adopted to connect elements with both upstream and downstream neighbors, considering possible surcharge flow and backflow effects. A convolution range threshold $\delta$ is designed to delineate the neighbors $\mathcal{N}_i$ of each element $i$ and formulate the binary adjacency matrixes $A_x = [a_{ij}]_{i,j \in \mathbf{V}}$ and $A_e = [a_{ij}]_{i,j \in \mathbf{E}}$ of nodes and edges based on the shortest drainage path distance (m) $\text{dist}(i, j)$ in the network as Figure 3(c) and Eq. 2-3. The threshold value $\delta$ of each GNN layer is determined to cover the flow routing distance in the prediction horizon after expansion in multiple layers.

$$\mathcal{N}_i = \{j | \text{dist}(i, j) \leq \delta\} \tag{2}$$

$$a_{ij} = \begin{cases} 1 & j \in \mathcal{N}_i \\ 0 & j \notin \mathcal{N}_i \end{cases} \tag{3}$$

Instead of full connections of all the elements' features, a GNN layer aggregates features of each node or edge from its neighbor elements with the adjacency matrix to consider the drainage structure. The graph attention (GAT) mechanism is adopted in each GNN layer which integrates the attention mechanism to learn the spatial dependencies of neighbors with different importance (Veličković et al., 2017). To consider the correlations between nodes and connected edges, the outputs for node and edge features are concatenated in each GAT layer as Eq. 4-7 and Figure 3(d),

$$X' = \text{GAT}(\text{Concat}(X, (W_{ex} \odot M + b_{ex}) \cdot f(E)), A_x) \tag{4}$$

$$E' = \text{GAT}(\text{Concat}(E, (W_{xe} \odot M^{\text{T}} + b_{xe}) \cdot f(X)), A_e) \tag{5}$$

$$M = [m_{ij}]_{i \in \mathbf{V}, j \in \mathbf{E}} \tag{6}$$

$$m_{ij} = \begin{cases} 1 & j \in \mathcal{N}_i^+ \\ -1 & j \in \mathcal{N}_i^- \\ 0 & else \end{cases} \tag{7}$$

where $X, E$ are hidden node and edge features before the layer, $X', E'$ are hidden node and edge states after the layer. $f(\cdot)$ is a one-layer perceptron. $W_{ex}$ and $W_{xe}$, $b_{ex}$ and $b_{xe}$ are trainable weights and biases of incidence matrixes $M$ and $M^{\text{T}}$ (which represent node-edge connections as Eq. 6-7) with the same shape. $\mathcal{N}_i^-$ and $\mathcal{N}_i^+$ are directly connected inflow and outflow edge sets of node $i$. $\odot$ and $\cdot$ represent Hadamard (element-wise) product and matrix (matmul) product respectively.

## 2.4 Two physics-guided mechanisms

To incorporate physics-guided domain knowledge and hydraulic constraints into data-driven surrogate modelling, a node-edge fusion modelling technique (Eq. 8-9) and a flooding determination mechanism (Eq. 10-12) are designed in the constraint block to regulate the surrogate model.

### 2.4.1 Node-edge fusion modelling

After spatial and temporal processing, hydraulic states are calculated through the constraint block to restrict the output of prediction model with physical constraints. The node-edge fusion modelling establishes the mass balance between node and edge flow variables as Eq. 8-9,

$$Q_{in,t}^i = \sum_{j \in \mathcal{N}_i^-} Q_t^j \tag{8}$$

$$Q_{out,t}^i = \sum_{j \in \mathcal{N}_i^+} Q_t^j \tag{9}$$

where $m_{ij}$ is from the incidence matrix $M$ in Eq. 6-7. In node-edge fusion modelling,

the inflow $Q_{in,t}^i$ and outflow $Q_{out,t}^i$ of node $i$ are calculated based on the flow prediction $Q_t^j$ of its directly connected pipes (upstream $\mathcal{N}_i^-$ and downstream $\mathcal{N}_i^+$) through the constraint block, as there may be more than one upstream or downstream pipes for each node. In contrast, the individual node-wise model directly predicts the inflow and outflow $Q_{in,t}^i, Q_{out,t}^i$ of each node with spatio-temporal layers. It should be noted that the fusion modelling can be incorporated into both NN-based and GNN-based models, as a physical guidance from the graph structure of the drainage network.

### 2.4.2 Flooding determination

The flooding volume is estimated with predicted flow variables at each manhole after prediction, and calculated as the flow balance error of each node as Eq. 10 (Palmitessa et al., 2022), which is used as the "balance" method in this work. As flooding only occurs at surcharge conditions, the balance method may mistake prediction error at unflooded nodes as flooding, and a determination method named "classification" is proposed to tell whether flooding occurs before flooding estimations as Eq. 11,

$$Q_{w,t}^i = \max\left(Q_{in,t}^i + r_t^i - Q_{out,t}^i, 0\right) \tag{10}$$

$$Q_{w,t}^i = \begin{cases} \max\left(Q_{in,t}^i + r_t^i - Q_{out,t}^i, 0\right) & if \; p_{f,t}^i > 0.5 \\ 0 & else \end{cases} \tag{11}$$

Where $p_{f,t}^i$ is an output of a sigmoid-activated binary classifier which decides whether flooding occurs at node $i$ given $t$. Eq. 10 only uses flow balance to calculate flooding, while Eq. 11 establishes a classifier for flooding determination.

### 2.5 Training

According to the output variables of the spatial and temporal layers, the surrogate model is trained with simulation data generated by a physics-based model to minimize the loss function including 3 parts: the prediction errors of node and edge states and the flooding determination loss as Eq. 12,

$$Loss = MSE_{node} + MSE_{edge} + BCE_{flood}$$

$$= \sum_{i,t} \frac{\left(x_t^i - \hat{x}_t^i\right)^2}{NT} + \sum_{j,t} \frac{\left(e_t^j - \hat{e}_t^j\right)^2}{CT} + \sum_{i,t} \frac{-p_{f,t}^i \log \hat{p}_{f,t}^i - (1-p_{f,t}^i) \log\left(1-\hat{p}_{f,t}^i\right)}{NT} \tag{12}$$

where $MSE$ and $BCE$ represent mean squared error and binary cross entropy respectively, $N, C, T$ are the number of nodes, edges and time steps, and variables with hat represent predicted values. All the variables are normalized into 0-1.

## 2.6 Evaluation metrics

The prediction performance is evaluated with root mean squared error (RMSE) as Eq. 13, where $x_t^j$ and $\hat{x}_t^j$ represent true and predicted values, $D$ represents the number of data points, and $T$ represents the number of steps in a horizon. As a flooding only occurs at surcharge conditions, precision and recall metrics are used to evaluate the flooding determination methods as Eq. 14-16. $TP$ is the number of actual flooding steps that are predicted as flooding, $TN$ is the number of non-flooding steps that are predicted as non-flooding, $FP$ is the number of non-flooding steps that are predicted as flooding, and $FN$ is the number of flooding steps that are predicted as non-flooding.

$$RMSE = \sqrt{\sum_{j,t} \frac{\left(x_t^j - \hat{x}_t^j\right)^2}{DT}} \tag{13}$$

$$Accuracy = \frac{TP+TN}{TP+FP+TN+FN} \tag{14}$$

$$Precision = \frac{TP}{TP+FP} \tag{15}$$

$$Recall = \frac{TP}{TP+FN} \tag{16}$$

## 2.7 Technical implementation

Storm water management model (SWMM) is used as the simulation toolkit, which is wrapped in the Python environment with pystorms and swmm-api for results analysis and training preparation (Pichler, 2022; Rimer et al., 2023; Rossman, 2015). Surrogate modelling is established using Keras and Spektral in Tensorflow (Developers, 2021; Grattarola and Alippi, 2020). All the experiments are carried out on a Nvidia RTX A5000 GPU (24GB) and 16 vCPU Intel(R) Xeon(R) Platinum 8350C CPU @ 2.60GHz . Technical details refer to our repository at https://github.com/Zhiyu014/GNN-UDS/.

## 3. Case Study and Results

This section implements the proposed surrogate model for real-time hydraulic prediction based on a stormwater network as a case study. The hydraulic prediction problem is investigated using surrogate models with different structures compared with the simulation results by SWMM. The predictive modelling results of NN-based and GNN-based surrogate models are compared in prediction accuracy (Section 3.2) and time consumption (Section 3.4), to examine the performance gain and cost of graph layers. The designed node-edge fusion modelling (Eq. 8-9) and the flooding determination mechanisms (Eq. 10-12) are also evaluated by comparing the results of models with or without these mechanisms in Section 3.3. All the prediction models use a 60-min lead time.

## 3.1 Case Study

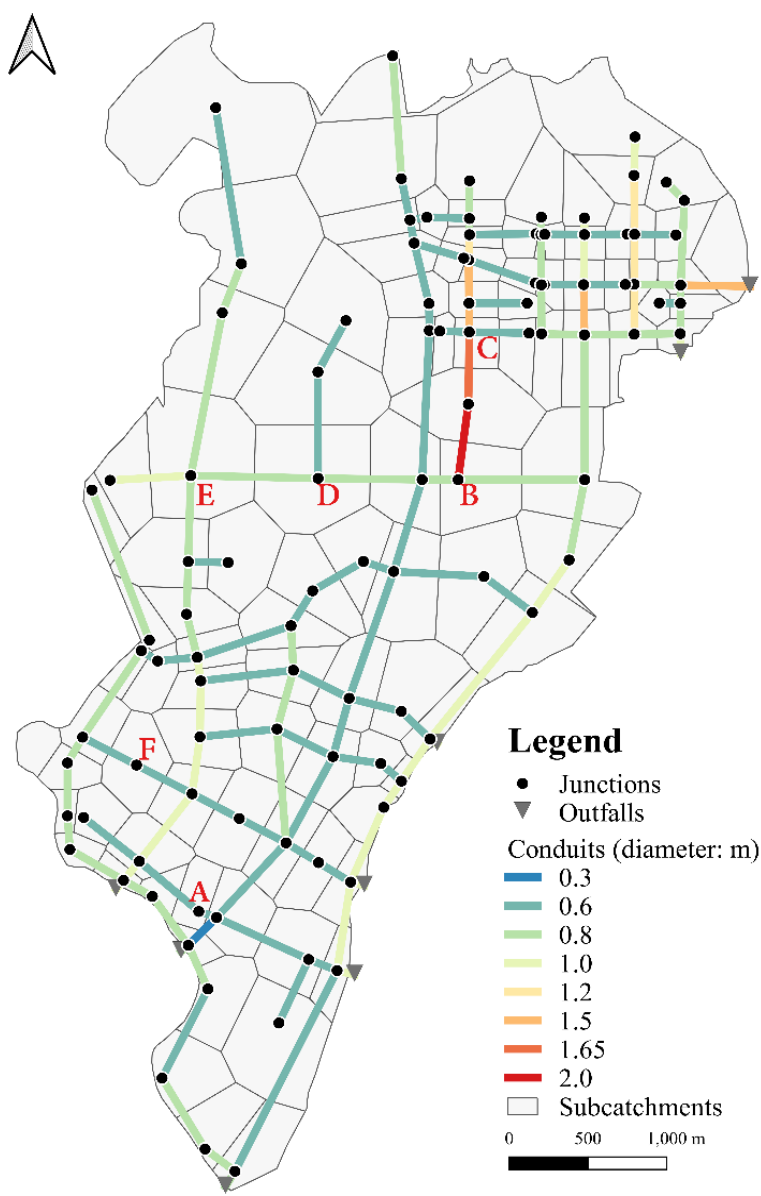

Figure 4. The Shunqing drainage network (Li et al., 2023) used in the case study.

A stormwater network Shunqing in Sichuan Province, China shown in Figure 4 (Li et al., 2023) is used to test the hydraulic prediction performance of the surrogate models. The Shunqing SWMM model has 105 manholes, 8 outfalls and 131 conduits. 148 rainfall events are provided in the model with durations of 6-24 hours. One hundred and eighteen (118) rainfall events are used to train the surrogate models, 27 used to validate the model, and the remaining 3, with 11-hour, 13-hour and 16-hour durations (R11, R13, R16), respectively, are used to test the model prediction performance.

The established prediction models are trained to predict the future 60-step states in a 60-min prediction horizon with the hydraulic states in the previous 60 steps and the runoff inflow in the prediction horizon as the input. The neighbor range threshold of the GNN-based model is set as $\delta = 1000\ m$ with 3 GNN layers in each spatial block to cover the flow distance of the 60-min prediction horizon. Since there is no controllable device for flow regulations in this case, the control boundary input $A_t$ is not considered in the surrogate model.

## 3.2 Spatial convolution performance

Since spatial lags need to be considered in flooding prediction (Li et al., 2023), a graph convolutional structure may also recognize the spatial correlation patterns of hydraulic states between neighbor elements (nodes and edges) efficiently. Therefore, an experiment is carried out to train the surrogate models with GNN and fully-connected NN as spatial convolution layer. GNN processes element features with respect to the inherent topological connections, whereas NN flattens all the elements and learns their correlations from scratch.

Figure 5 shows the flooding predictions at 6 junctions (see Figure 4 and also Li et al., 2023) by GNN-based and NN-based models, and the 2 physics-guided constraints are considered in both models. The trained GNN-based and NN-based models can provide close hydraulic predictions to that of SWMM, while the root mean squared error (RMSE) values of GNN are smaller than those of NN in all 3 rainfalls. The GNN model outperforms NN at all locations in capturing the flooding variations. In general, the GNN-based surrogate model has better performance than the model with fully-connected NN in the hydraulic predictions. The RMSE values of other hydraulic variables are summarized in Figure 6(a), whereas the full results of all the nodes and edges refer to the supplementary materials. Note that the timeseries plots (Figure 5 and those in the supplementary materials) only display the value in the last step of each prediction horizon (i.e., a 60-min lead time), but the RMSE values are averaged in the entire horizon.

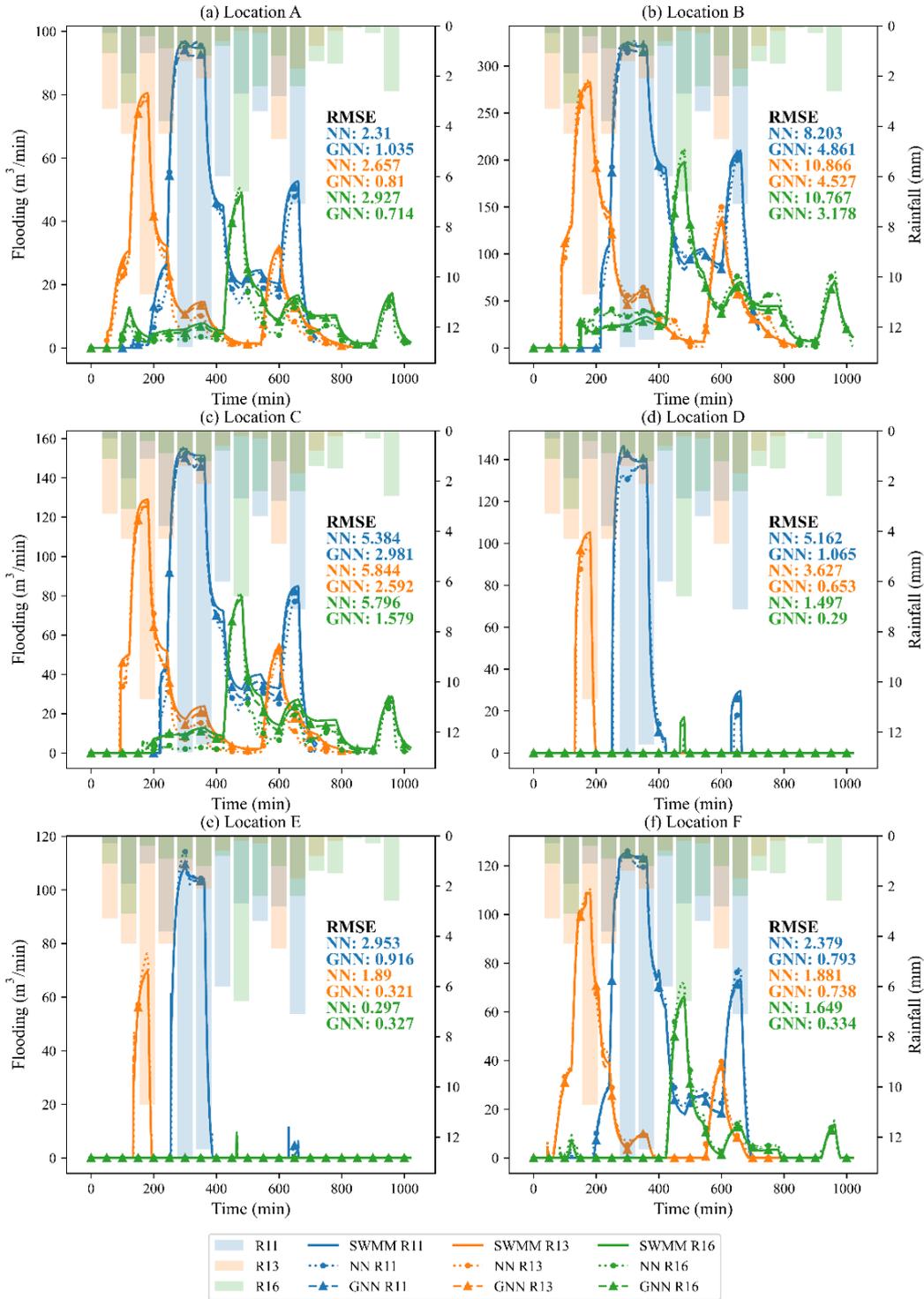

Figure 5. Flooding prediction results with 60-min lead time with the GNN and fully-connected neural network (NN) models at 6 representative junctions (A-F, Figure 5) and 3 testing rainfalls.

### 3.3 Effects of physics-guided mechanisms

A node-edge fusion modelling technique (Eq. 8-9) and flooding determination methods (Eq. 10-12) are designed as constraints in the surrogate model, and this section analyzes their performance gain in hydraulic prediction. The node-edge fusion is incorporated in both the NN-based and GNN-based models to formulate 4 prediction methods including NN-individual, NN-fusion, GNN-individual, and GNN-fusion, which are tested for multi-variable predictions. A flooding determination method is implemented in each of the 4 prediction methods for node flooding estimation compared with the balance method.

### 3.3.1 Node-edge fusion modelling

Figure 6(a) summarizes the prediction errors (RMSE) in 5 hydraulic variables by the NN and GNN models with and without node-edge fusion. According to the results of GNN-individual and GNN-fusion, the node-edge fusion model outperforms the individual model in flooding prediction, which highly relies on the prediction accuracy of node inflow and outflow. Since the fusion model calculates node inflow and outflow with edge flow in Eq. 8-9 as physical constraints, there are improvements in node flow variables for better flooding estimations than direct predictions from the model. There is small deterioration in edge flow as shown in Figure 6(a) due to the highlight of node flow loss with fusion modelling in the training process. The errors of NN-individual and NN-fusion in Figure 6 indicate that node-edge fusion modelling also significantly reduces prediction errors of the NN-based model, which has no knowledge of drainage element connections in the spatio-temporal blocks, and is informed of the physical relationship between connected flow variables by the constraint.

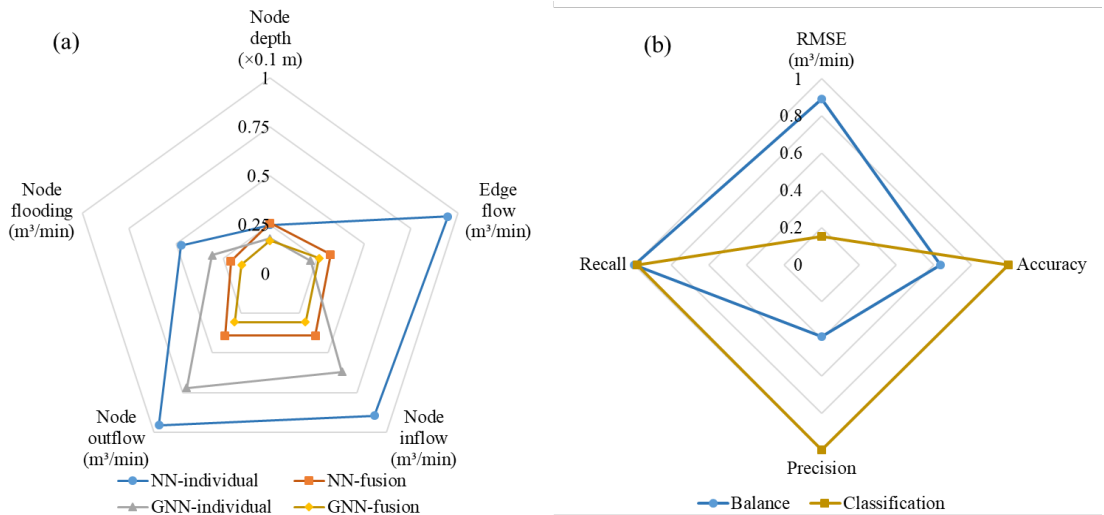

Figure 6. (a) Root mean squared errors of predictions in 5 hydraulic variables by the GNN and fully-connected NN models with and without the node-edge fusion mechanism. (b) Flooding estimation errors (RMSE) and determination metrics of balance and classification methods based on the GNN model with node-edge fusion.

### 3.3.2 Flooding determination

Figure 6(b) shows the performance gain of the classification method in estimating node flooding and determining flooding steps compared with the balance, using the GNN-based fusion model. A classifier (Eq. 11) can determine whether flooding occurs, and provide more accurate flooding estimations than the balance method (Eq. 10), which tends to overestimate the flooding volume in each step. Low precision and accuracy values indicate that the balance method still calculates flooding even there is no surcharge at a step, and it does not miss any flooding step with a high recall value. Since flow routing errors may exist at any node and time step, directly using flow balance to calculate flooding volume can be biased. A trained classifier determines surcharge steps with high accuracy and precision, as well as a slightly lower recall than the balance method, and effectively solves the surcharge condition problem in node flooding estimation.

### 3.4 Time consumption

The time consumption of the prediction model is important for real-time applications which require instant feedback for online analysis or optimization. A 60-min predictive modelling scenario is used to examine the efficiency of the physics-based model (i.e. the SWMM model) and the surrogate model. The computational time results are shown in Table 1 where the surrogate model consumes around 1% of the time by the SWMM in a single simulation run. Real-time modelling often requires repetitive simulations of future scenarios with different boundary conditions, and the computing efficiency gain of the surrogate models becomes more significant in 32 simulations as shown in Table 1 due to GPU parallelization, compared with CPU multiprocessing for SWMM (Sadler et al., 2019). As shown in Table 1, at the costs of several hours to train a surrogate model for 20k epochs, the trained models save much computational time for online analyses compared to the physics-based model, rendering these models more suitable for real-time applications.

Table 1. Computational time for model training and 60-min predictive simulations using the SWMM and surrogate models.

| | SWMM | NN-based surrogate | GAT-based surrogate |
|---|---|---|---|
| Training (20k epochs) | - | 0.71 h | 2.14 h |
| 60-min predictive simulation | 5.70 s | 0.034 s | 0.064 s |
| 60-min predictive simulation (32 times) | 21.44 s | 0.094 s | 0.145 s |

Note: Predictive simulation time results are averaged across all the time steps in 3 testing rainfall events.

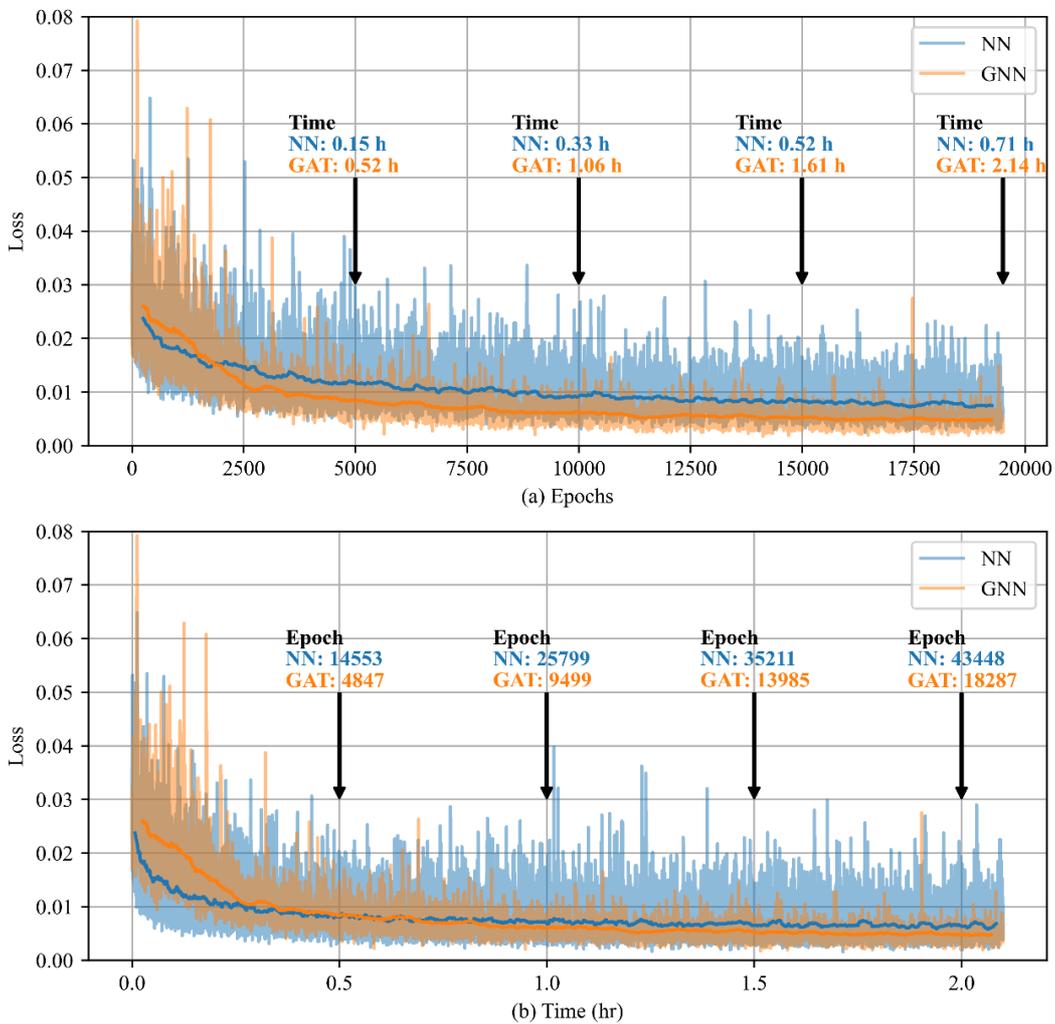

Figure 7. Validation losses in the training process of the GNN and fully-connected NN surrogate models with respect to training epochs (a) and time (b). Note that the solid lines represent moving average losses over 500 epochs.

As shown in Figure 7(a), the GNN-based model achieves lower losses and converge faster than the NN-based model in training epochs. The GNN-based surrogate model appears to require longer training and prediction time than the NN-based one. This is caused by the fact that GAT increases the model complexity with more trainable variables than the NN-based model. However, the GNN-based model keeps lower training losses and more stable than the NN-based one after 0.5 h (Figure 7(b)), and is therefore a more cost-effective surrogate model.

## 4. Discussion

### 4.1 Spatial convolution structure

The proposed GNN-based structure is a cost-effective and interpretable solution for hydraulic predictive modelling in urban drainage networks. Differing from fully-connected layers, it considers physical topology connections and aggregates neighbor information along the drainage network for enhancing the modelling performance.

Figure 5 and Figure 6(a) show that the GNN layers outperform fully-connected NN layers in predicting hydraulic states like water depth, flow, and flooding, after equal training epochs. The flow routing process adheres to the topology connections of the drainage network, and is represented as hydraulic variables of manholes and conduits in surrogate modelling. Fully-connected layers in the NN-based model flatten all elements' features (Figure 3(a)) and derive their relationship and interactions from scratch in the training process, which can be quite difficult without the existing drainage topology as a reference. As graph layers are designed to handle graph-represented data (Figure 3(c)) by leveraging inductive message passing and aggregation along the topology structure, stacking graph layers in neural networks could mimic flow routing along connected manholes. Therefore, the GNN-based model achieves more accurate predictions than the NN-based one in multiple hydraulic variables (especially node depth and edge flow) (Figure 6), which particularly depend on upstream and downstream dynamics.

Graph layers process data features via neighbor convolution to avoid unnecessary information exchange, while fully-connected layers may face dimensionality issues with large drainage networks (Garzón et al., 2022). For relatively simple problems with less than 100 junctions and conduits, it may be feasible for a fully-connected model with hundreds of neural channels to learn from the hydraulic dynamics of all of them to simulate the water depth and flow variables. If the drainage network scale or the prediction horizon expands, a much larger fully-connected model is in need to process the widened features in Figure 3(b) as the problem complexity increases, or the fitting performance would decay. As shown in Figure 3(c)(d), GNN provides an inductive bias for element-level prediction, in which a filter based on adjacency matrix allocates a relevant neighborhood of each node or edge (Eq. 2-3) for feature aggregation with attention to focus on the relevant interactions and avoid the dimensionality problem. Therefore, the GNN-based model simplifies the modelling problem with inherent relationship of each center node with its connected nodes and edges. This helps to achieve lower training losses than the NN-based model, as shown in Figure 7, making it a cost-effective solution for hydraulic prediction.

The model structure has a large impact on its interpretability, and blindly feeding all the data into a fully-connected NN is not a wise solution to such a specific problem in urban drainage networks. The general idea (Figure 2) is to imitate the spatio-temporal dynamics of hydraulic mechanisms in the drainage network through two sets of spatial and temporal blocks with multiple graph layers in both past and future domains, driven by external runoff inflow and control actions of pumps or valves (if any) as boundaries. Different from step-by-step simulation in Palmitessa et al. (2022), this model follows the timeseries paradigm (Jin et al., 2023b) as a one-step state transition model in a certain horizon (60 minutes in the case study) for real-time use, which can be further used as the internal model of model predictive control (Lund et al., 2018).

## 4.2 Physics-guided mechanisms in data-driven modelling

Data-driven modelling shows great power in various problems, but physics-guided learning (Feng et al., 2022; Ye et al., 2022) or data-physics coupled models (Ye et al., 2023) could better fit in the application problem with enhanced performance and stronger interpretability (Garzón et al., 2022; Vidyarthi et al., 2023) than pure black-box models. As deep learning models are trained to map the data distributions from input to output, there are uncertainties and errors in prediction variables, while physics-guided mechanisms could form parts of the model or restrict the prediction output with hydraulic constraints to limit the error accumulations. In urban drainage modelling, the flow balance between connected elements and flooding conditions can be incorporated as node-edge fusion modelling (Eq. 8-9) and flooding determination (Eq. 10-12) mechanisms to consider the constraints and relationship between hydraulic variables. Since flooding is a key variable in real-time drainage modelling (Li et al., 2023; Tian et al., 2022), its calculation relies on two essential parts including node flow predictions and node surcharge judgments. According to Figure 6 and Table 1, the two mechanisms incorporated could increase the prediction performance of flow variables and determine node surcharge conditions, respectively, to limit flooding estimation errors.

Node-edge fusion modelling reduces node flow and flooding prediction errors compared with the individual node-wise models (Figure 6). An individual node-wise model predicts node inflow and outflow variables independently, but ignores existing physical relationship between model outputs and provides biased results. In contrast, node-edge fusion modelling connects the outputs of node and edge flow variables to establish flow balance constraints (Eq. 8-9) in both training and application stages. This effectively limits the large prediction errors in node inflow and outflow with edge flow predictions, which are important in node flooding estimation.

As flooding only exists at surcharged nodes, it is important to determine accurately when and where flooding occurs for accurate predictions. Flooding can be regarded as flow balance error at junctions in surcharge conditions (Palmitessa et al., 2022). The balance (Eq. 10) performance in Figure 6 reveals that directly using flow error at each junction and step (Eq. 10) results in overestimations including flow routing errors at unflooded conditions. Therefore, a judgment criterion is in need to determine whether flooding occur. A flexible way is to treat the surcharge condition determination as a binary classification problem, and establish a classifier based on the neural network to output flooding occurrence probabilities as Eq. 11, taking hydraulic states and prior conditions into account. Figure 6 demonstrates that the classification method enhances flooding estimation performance compared with the balance method, after training with flooding occurrence scenarios.

Figure 6 indicates that the node-edge fusion modelling also significantly increases the prediction performance in both flow and water depth variables of the NN-based model, which is informed of drainage topology by the fusion constraint and regulated by node-edge flow balance to consider the graph structure in surrogate training, similar to the idea of GNN. However, the NN-based models still cannot overtake GNN-based models in multi-variable predictions as shown in Figure 6.

### 4.3 Modelling efficiency

The physics-based model has large computational burden for solving complex partial differential equations, while a surrogate model only calculates feedforward through the neural network in real time, and a learning-based surrogate also exploits parallel computing acceleration of multiple simulations in a graph processing unit (GPU). The trade-off exists in modelling efficiency and complexity, while it is necessary for a surrogate model to preserve accuracy and interpretability at the cost of efficiency loss.

In model predictive control (MPC) scenarios, a predictive model is needed to evaluate massive possible control policies in the action space of multiple actuators for real-time decision-making (Lund et al., 2018), and the number of solutions increase dramatically if each valve/orifice has various opening settings or each pumping station has multiple pumps. Surrogate modelling can accelerate the online simulation for real-time optimization use (Garzón et al., 2022), which requires second-level or faster simulations, and the low efficiency of physics-based models restricts their feasibility. Table 1 demonstrates that a physics-based model is inefficient for 60-min online predictive modelling with around 20 s for one iteration, as the computational time is out of control after iterative simulation-optimization, leading to a long delay of minutes in each MPC step. Therefore, a surrogate model can significantly alleviate the online computational burden of control evaluations (Luo et al., 2023), as both the NN-based and GNN-based models shorten the computational time of 32 simulations to less than one second through parallel computing in a batch with GPU acceleration in Table 1. Although the physics-based model is also parallelized with multiprocessing in CPU (Sadler et al., 2019), its acceleration performance is less than the data-driven surrogate.

Although the proposed GNN-based model consumes slightly longer training and modelling time than the NN-based model as shown in Table 1, it achieves lower and more stable losses in the training process with respect to both training epochs and time (Figure 7), and shows higher learning efficiency and better generalizability in different validation datasets. There is a trade-off between the model complexity and computation speed. Purely data-driven models such as fully-connected NN models are faster and simpler but are black-boxes and hence have poorer interpretability. At the cost of computation deceleration, the GNN-based model preserves high accuracy in real-time hydraulic predictions (Figure 5), and also increases the interpretability with inductive biases from physical drainage topology, as a cost-effective physics-guided data-driven model of urban drainage networks (Palmitessa et al., 2022).

## 5. Conclusions

This work proposes a surrogate flow routing model for real-time hydraulic predictions in urban drainage networks based on the graph neural network (GNN) structure. The proposed approach also incorporates two physics-guided mechanisms as physical constraints into the surrogate model. The model is trained and tested, with hydraulic prediction performance compared with simulation results obtained with a physics-based model in a stormwater network case. According to the hydraulic prediction results, three main conclusions can be drawn.

1) The GNN-based model is more effective than the fully-connected NN-based model in element-level hydraulic predictions of urban drainage networks after equal training epochs. Derived from topological neighbor information, GNN leverages inductive bias for the model to follow flow routing pattern along the drainage network, and enables a cost-effective surrogate model for real-time hydraulic prediction.

2) Physics-guided mechanisms limit the model prediction error with hydraulic constraints and achieve better interpretability than purely data-driven black-box models. Node-edge fusion enhances prediction performance of flow variables with flow constraints between connected manholes and conduits than the individual node-wise model. The flooding determination method determines whether flooding occurs with a classifier to limit flooding volume estimations in surcharge conditions and exclude flow balance error in unflooded conditions.

3) The surrogate models accelerate hydraulic predictions for real-time predictive modelling scenarios, as they avoid recursive equation solving of physics-based models and enables parallel computing in graph processing units. The GNN-based model costs slightly higher computational load than the NN-based one but preserves better prediction performance and higher interpretability.

In summary, the proposed GNN-based surrogate model can accurately and efficiently predict hydraulic variations with past states and future boundary conditions in the drainage network, and also preserves interpretability with designed physics-guided mechanisms to incorporate hydraulic constraints into data-driven modelling. The GNN-based surrogate model can be further exploited to generalize different objective functions of learning-based model predictive control.

## Acknowledgments

This work was supported by the National Natural Science Foundation of China

(Grant No. 52170102). The authors would like to appreciate Prof. Chunxiao Zhang for providing the Shunqing model (https://github.com/lhmygis/ga_ann_for_uds) in open access. Zhiguo Yuan is a Global STEM Professor jointly funded by the Innovation, Technology and Industry Bureau ("ITIB") and Education Bureau ("EDB") of the Government of the Hong Kong Special Administrative Region, and acknowledges financial support from the Hong Kong Jockey Club for the JC STEM Lab of Sustainable Urban Water Management.

**Data availability**

The source code, data, and related files can be found in the repository (https://github.com/Zhiyu014/GNN-UDS).